\documentclass[10pt,twocolumn,letterpaper]{article}

\usepackage{iccv}
\usepackage{times}
\usepackage{epsfig}
\usepackage{graphicx}
\usepackage{amsmath,bm}
\usepackage{amssymb}

\usepackage{graphicx}
\newcommand{\cev}[1]{\reflectbox{\ensuremath{\vec{\reflectbox{\ensuremath{#1}}}}}}

\usepackage[pagebackref=true,breaklinks=true,letterpaper=true,colorlinks,bookmarks=false]{hyperref}

\usepackage[breaklinks=true,bookmarks=false]{hyperref}

\iccvfinalcopy 

\begin{document}
\pagenumbering{gobble}

\title{Spatiotemporal Modeling for Crowd Counting in Videos}

\author{Feng Xiong ~ Xingjian Shi ~ Dit-Yan Yeung\\
	Department of Computer Science and Engineering \\
	Hong Kong University of Science and Technology\\
	{\tt\small \{fxiongab,xshiab,dyyeung\}@cse.ust.hk}
}

\maketitle

\begin{abstract}
	Region of Interest (ROI) crowd counting can be formulated as a regression problem of learning a mapping from an image or a video frame to a crowd density map.  Recently, convolutional neural network (CNN) models have achieved promising results for crowd counting.  However, even when dealing with video data, CNN-based methods still consider each video frame independently, ignoring the strong temporal correlation between neighboring frames.  To exploit the otherwise very useful temporal information in video sequences, we propose a variant of a recent deep learning model called convolutional LSTM (\mbox{ConvLSTM}) for crowd counting.  Unlike the previous CNN-based methods, our method fully captures both spatial and temporal dependencies.  Furthermore, we extend the ConvLSTM model to a bidirectional ConvLSTM model which can access long-range information in both directions.  Extensive experiments using four publicly available datasets demonstrate the reliability of our approach and the effectiveness of incorporating temporal information to boost the accuracy of crowd counting. In addition, we also conduct some transfer learning experiments to show that once our model is trained on one dataset, its learning experience can be transferred easily to a new dataset which consists of only very few video frames for model adaptation.
\end{abstract}

\section{Introduction}

Crowd counting is the problem of estimating the number of people in a still image or a video.  It has drawn a lot of attention due to the need for solving this problem in many real-world applications such as video surveillance, traffic control, and emergency management.  Proper use of crowd counting techniques can help to prevent some serious accidents such as the massive stampede that happened in Shanghai, China during the 2015 New Year's Eve festivities, killing 35 people.  Moreover, some crowd counting methods can also be applied to other object counting applications such as cell counting in microscopic images~\cite{densitymap,cnnboosting}, vehicle counting in public areas~\cite{vehiclecounting,vehiclecoutning2}, and animal counting in the wild~\cite{wildcounting}.

The methods for crowd counting in videos fall into two broad categories: (a)~Region of Interest (ROI) counting, which estimates the total number of people in some region at a certain time; and (b)~Line of Interest (LOI) counting, which counts the number of people crossing a detecting line in a video during a certain period of time.  Since LOI counting is more restrictive in its applications and is much less studied than ROI counting, we focus on ROI counting in this paper.

Many methods have been proposed in the past for crowd counting.  Some methods take the approach of tackling the crowd counting problem in an unsupervised manner through grouping based on self-similarities~\cite{extract} or motion similarities~\cite{motions}.  However, the accuracy of such fully unsupervised counting methods is limited. Thus more attention has been paid to the supervised learning approach.  Supervised crowd counting methods generally fall into two categories: detection-based methods and regression-based methods.  In detection-based methods, some given object detectors~\cite{detect1,detect2,detect3,detect4} are used to detect people individually.  They usually operate in two stages by first producing a real-valued confidence map and then locating from the map those peaks that correspond to individual people.  Once the locations of all individuals have been estimated, the counting problem becomes trivial.  However, object detection could be a challenging problem especially under severe occlusion in highly crowded scenes.

In recent years, regression-based methods have achieved promising counting results in crowded scenes.  Regression-based methods avoid solving the difficult detection problem.  Instead, they regard crowd counting as a regression problem by learning a regression function or mapping from some holistic or local visual features to a crowd count or a crowd density map.  Linear regression, Gaussian process regression, and neural networks are often used as the regression models. Currently, most methods which achieve state-of-the-art performance are regression-based methods~\cite{gpr,ridge,interactive,rf,cum,dcnn,mcnn,cnnboosting}.

With the recent resurgence of interest in convolutional neural network~(CNN) models which have reported promising results for many computer vision tasks~\cite{krizhevsky2012imagenet}, in the recent two years some CNN-based methods~\cite{dcnn,mcnn,cnnboosting,workshop} have also been proposed for crowd counting, giving state-of-the-art results on the existing crowd counting datasets such as UCSD~\cite{gpr} and UCF~\cite{ucf50}.  Unlike traditional regression-based methods~\cite{gpr,ridge}, CNN-based methods do not need handcrafted features but can learn powerful features in an end-to-end manner.  However, even when dealing with video datasets, these CNN-based methods still regard the data as individual still images and ignore the strong temporal correlation between neighboring video frames.

In this paper, we propose a variant of a recent deep learning model called convolutional LSTM (ConvLSTM)~\cite{convlstm} for crowd counting. While CNN-based methods exploit only spatial features by ignoring the otherwise very useful temporal information in video sequences, our method fully captures both spatial and temporal dependencies.  Incorporating the temporal dimension is important as it is well known that motion information can help to boost the counting accuracy in complex scenes. Thorough experimental validation using four publicly available datasets will be reported later in this paper to demonstrate the effectiveness of incorporating temporal information to boost the accuracy of crowd counting.

\section{Related Work}

\subsection{Deep learning methods for crowd counting}

C.~Zhang \emph{et al.}~\cite{dcnn} proposed the first CNN-based method for crowd counting. Following this work, Y.~Zhang \emph{et al.} \cite{mcnn} proposed a multi-column CNN architecture which allows the input image to be of arbitrary size or resolution.  The multi-column CNN architecture also uses a different method for computing the crowd density.  Walach and Wolf~\cite{cnnboosting} proposed a stage-wise approach by carrying out model training in stages.  In the spirit of the gradient boosting approach, CNNs are added one at a time such that every new CNN is trained to estimate the residual error of the earlier prediction.  After the first CNN has been trained, the second CNN is trained on the difference between the current estimate and the learning target.  The third CNN is then added and the process continues.
Rubio \emph{et al.}~\cite{hydracnn} proposed a framework called Hydra CNN which uses a pyramid of image patches extracted at multiple scales to perform the final density prediction.  All these methods have reported good results for the UCSD dataset. However, to the best of our knowledge, temporal dependencies have not been explicitly exploited by deep learning models for crowd counting.  These CNN-based methods simply treat the video sequences in the UCSD dataset as a set of still images without considering their temporal dependencies.

\subsection{Density map regression for crowd counting}

Lempitskey and Zisserman~\cite{densitymap} proposed a method to change the target of regression from a single crowd count to a crowd density map. We note that crowd density is more informative than crowd count, since the former also includes location information of the crowd.  With a crowd density map, the crowd count of any given region can be estimated easily.  The crowd count of the whole image is simply the integral of the density function over the entire image. All CNN-based methods mentioned above have used the crowd density map as the regression target.

\subsection{ConvLSTM for spatiotemporal modeling}

Recurrent neural networks (RNNs) have been applied successfully to various sequence learning tasks~\cite{lstm}. The incorporation of long short-term memory~(LSTM) cells enables RNNs to exploit longer-term temporal dependencies. By extending the fully connected LSTM (FC-LSTM) to have convolutional structures in both the input-to-state and state-to-state connections, Shi \emph{et al.}~\cite{convlstm} proposed the \mbox{ConvLSTM} model for precipitation nowcasting which is a spatiotemporal forecasting problem. The ConvLSTM layer not only preserves the advantages of FC-LSTM but is also suitable for spatiotemporal data due to its inherent convolutional structures.

ConvLSTM models have also proven effective for some other spatiotemporal tasks. Finn \emph{et al.}~\cite{finn2016unsupervised} employed stacked ConvLSTMs to generate motion predictions. Villegas \emph{et al.}~\cite{iclr} proposed a ConvLSTM-based method to model the spatiotemporal dynamics for pixel-level prediction in natural videos. Also, Y.~Zhang \emph{et al.}~\cite{zhang2016} applied network-in-network principles, batch normalization, residual connections, and ConvLSTMs to build very deep recurrent and convolutional structures for speech recognition.

\section{Our Crowd Counting Method}

\subsection{Crowd density map}

Following the previous work~\cite{densitymap} as reviewed above, we also formulate crowd counting as a density map estimation problem.  Compared to methods that give an estimated crowd count of the whole image as output, methods that give a crowd density map also provide location information about the crowd distribution which is useful for many applications.

We assume that each training image $I_{i}$ is annotated with a set of 2D points $\mathcal{P}_{i} = { \{ P_{1}, \ldots, P_{C(i)} \} }$, where $C(i)$ is the total number of people annotated.  We define the ground-truth density map for supervised learning as a sum of Gaussian kernels each of which is centered at the location of one person.
The ground-truth density map $F_{i}(p)$ for image $I_{i}$ can be defined as follows:
\begin{equation}
	\forall p \in I_{i}, \ F_{i}(p) = \sum_{P\in \mathcal{P}_{i}} \mathcal{N}(p; P,\sigma^{2}I_{2\times2}),
\end{equation}
where $p$ denotes a pixel in image $I_i$, $\mathcal{P}_{i}$ is the set of annotated points (usually corresponding to the positions of the human heads), $ \mathcal{N}(p; P,\sigma^{2}I_{2\times2}) $ represents a normalized 2D Gaussian kernel evaluated at the pixel position $p$ with its mean at the head position $P$ and an isotropic $2 \times 2$ covariance matrix $I_{2\times2}$ with variance $\sigma^2$.

For annotated points which are close to the image boundary, part of their probability mass will reside outside the image.  Consequently, integrating the ground-truth density map over the entire image will not match the crowd count exactly.  Fortunately, this effect can be neglected for most applications because the differences are generally small. Moreover, in many cases, a pedestrian who lies partially outside the image boundary should not be counted as a whole person.

Another subtlety that is worth noticing is that the images are often not captured with a bird's-eye view and hence leads to perspective distortion.  As a result, the pixels associated with different annotated points correspond to regions of different scales in the actual 3D scene.  To overcome the effects due to perspective distortion, we need to normalize the crowd density map with the perspective map $M(p)$. The pixel value in the perspective map represents the number of pixels in the image corresponding to one meter at that location in the real scene. In our experiments, we set $ \sigma = 0.3 M(p)$ and then normalize the whole distribution to ensure that the sum of ground-truth density is equal to the crowd count of the image.

\subsection{ConvLSTM model}

FC-LSTM has proven powerful for handling temporal correlations, but it fails to maintain structural locality. To exploit temporal correlations for video crowd counting, we propose a model based on ConvLSTM~\cite{convlstm} to learn a density map.  As an extension of FC-LSTM, ConvLSTM has convolutional structures in both the input-to-state and state-to-state connections.
We can regard all the inputs, cell outputs, hidden states $\mathcal{H}_{1},...,\mathcal{H}_{t}$, and gates $i_{t}, f_{t}, o_{t}$ of the ConvLSTM as 3D tensors whose last two dimensions are spatial dimensions. The outputs of ConvLSTM cells depend on the inputs and past states of the local neighbors. The key equations of ConvLSTM are shown in (\ref{eq:1}) below, where `*' denotes the convolution operator, `$\circ$' denotes the Hadamard product, and $\sigma(\cdot)$ denotes the logistic sigmoid function:
\begin{equation} \label{eq:1}
	\begin{split}
		i_{t} &= \sigma(W_{xi}*\mathcal{X}_{t} + W_{hi}*\mathcal{H}_{t-1} + W_{ci}\circ \mathcal{C}_{t-1} + b_{i}),   \\
		f_{t} &= \sigma(W_{xf}*\mathcal{X}_{t} + W_{hf}*\mathcal{H}_{t-1} + W_{cf}\circ \mathcal{C}_{t-1} + b_{f}),   \\
		\mathcal{C}_{t} &= f_{t}\circ \mathcal{C}_{t-1} + i_{t}\circ \tanh(W_{xc}*\mathcal{X}_{t} + W_{hc}*\mathcal{H}_{t-1} + b_{c}),  \\
		o_{t} &= \sigma((W_{xo}*\mathcal{X}_{t} + W_{ho}*\mathcal{H}_{t-1} + W_{co}\circ \mathcal{C}_{t} + b_{o}), \\
		\mathcal{H}_{t} &= o_{t} \circ \tanh(\mathcal{C}_{t}).
	\end{split}
\end{equation}

\begin{figure}
	\centering
	\includegraphics[width = 0.45 \textwidth]{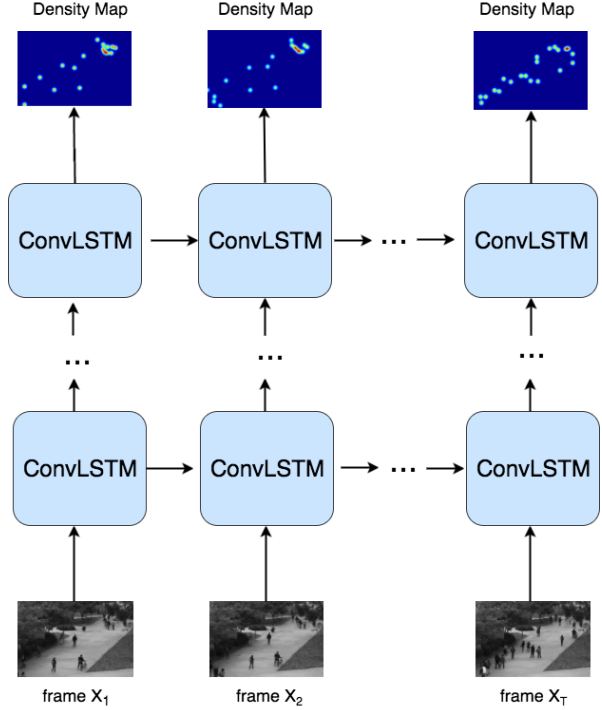}
	\caption{ConvLSTM model for crowd counting}
	\label{fig:1}
\end{figure}

Figure~\ref{fig:1} shows our ConvLSTM model for crowd counting where each building block involves a ConvLSTM.

The inputs $\bm{\mathcal{X}}_{1:t} = \mathcal{X}_{1}, \ldots, \mathcal{X}_{t}$ are consecutive frames of a video and the cell outputs $ \mathcal{C}_{1}, \ldots, \mathcal{C}_{t}$ are the estimated density maps of the corresponding frames.
If we remove the connections between ConvLSTM cells, we can regard each ConvLSTM cell as a CNN model with gates. We set all the input-to-state and state-to-state kernels to size $5 \times 5$ and the number of layers to 4. To relate the feature maps to the density map, we adopt filters all of size $1 \times 1$. We use the Euclidean distance to measure the difference between the estimated and ground-truth density maps. So we define the loss function $L(\theta)$ between the estimated density map $F(\bm{\mathcal{X}}_{1:t}; \theta)$ and the ground-truth density map $D_{t}$ as follows:
\begin{equation}
	L(\theta) = \frac{1}{2T} \sum_{t=1}^{T} \left \| F(\bm{\mathcal{X}}_{1:t}; \theta) - D_{t} \right \|_{2}^{2},
\end{equation}
where $T$ is the length of the video clip and $\theta$ denotes the parameter vector.

\subsection{From ConvLSTM to bidirectional ConvLSTM}

\begin{figure}
	\centering
	\includegraphics[width = 0.45 \textwidth]{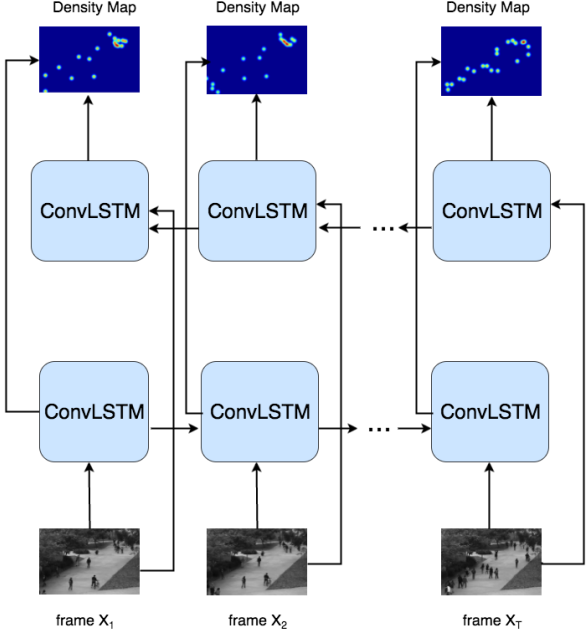}
	\caption{Bidirectional ConvLSTM model for crowd counting}
	\label{fig:2}
\end{figure}
Inspired by \cite{brcn,zhang2016}, we further extend the ConvLSTM model to a bidirectional ConvLSTM model which can access long-range information in both directions.

Figure~\ref{fig:2} depicts the bidirectional ConvLSTM model for crowd counting.  Its inputs and outputs are the same as those in the ConvLSTM model.  It works by computing the forward hidden sequence $ \vec{h}$, backward hidden sequence $ \cev{h} $, and output sequence by iterating backward from $ t = T$ to $t=1$, iterating forward from $ t = 1$ to $t = T$, and then updating the output layer. If we denote the state updating function in~\eqref{eq:1} as $\mathcal{H}_t, \mathcal{C}_t = \text{ConvLSTM}(\mathcal{X}_t, \mathcal{H}_{t-1}, \mathcal{C}_{t-1})$, the equation of bidirectional ConvLSTM can be written as follows:
\begin{equation}
	\begin{aligned}
		\vec{\mathcal{H}}_{t}, \vec{\mathcal{C}}_{t} &= \text{ConvLSTM}(\mathcal{X}_{t}, \vec{\mathcal{H}}_{t-1}, \vec{\mathcal{C}}_{t-1}),   \\
		\cev{\mathcal{H}}_{t}, \cev{\mathcal{C}}_{t} &= \text{ConvLSTM}(\mathcal{X}_{t}, \cev{\mathcal{H}}_{t+1}, \cev{\mathcal{C}}_{t+1}),   \\
		\mathcal{Y}_{t} &= \text{concat}(\vec{\mathcal{H}}_{t}, \cev{\mathcal{H}}_{t}),
	\end{aligned}
	\label{eq:bconvlstm}
\end{equation}
where $\mathcal{Y}_t$ is the output at timestamp $t$.

Y.~Zhang \emph{et al.}~\cite{zhang2016} found that bidirectional ConvLSTM consistently outperforms its unidirectional counterpart in speech recognition. In the next section, we also compare bidirectional ConvLSTM with the original ConvLSTM for crowd counting using different datasets.  

\subsection{ConvLSTM-nt: a degenerate variant of ConvLSTM for comparison }

\begin{figure}
	\centering
	\includegraphics[width = 0.45 \textwidth]{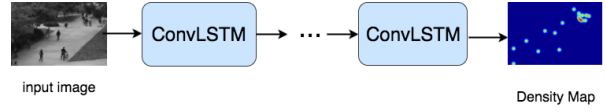}
	\caption{ConvLSTM-nt model for crowd counting}
	\label{fig:0}
\end{figure}

To better understand the effectiveness of exploiting temporal information, we propose a degenerate variant of ConvLSTM, called ConvLSTM with no temporal information (ConvLSTM-nt), by removing all connections between the ConvLSTM cells. ConvLSTM-nt can be seen as a CNN model with gates. The parameters of ConvLSTM-nt are the same as those of ConvLSTM introduced above. The structure of ConvLSTM-nt is shown in Figure~\ref{fig:0}. 

All our three models have 4 layers, with $128,64,64 $ and $64$ hidden states respectively in the four ConvLSTM layers. For the training scheme, we train all models using the TensorFlow library, optimizing to convergence using ADAM~\cite{ADAM} with the suggested hyperparameters in TensorFlow.

In the experiments to be reported in the next section, whenever the dataset consists of still images not forming video sequences, both the original ConvLSTM and our bidirectional extension cannot be used but only ConvLSTM-nt will be used.	

\section{Experiments}

We conduct comparative study using four annotated datasets which include the UCF\_CC\_50 dataset~\cite{ucf50}, UCSD dataset~\cite{gpr}, Mall dataset~\cite{ridge}, and WorldExpo'10 dataset~\cite{dcnn,zhang2016data}. Some statistics of these datasets are summarized in Table \ref{tab:all}. We also conduct experiments in the transfer learning setting by using one of the UCSD and Mall datasets as the source domain and the other one as the target domain.
\begin{table*}[h]
	\begin{center}
		\caption{Statistics of the four datasets \newline}
		\label{tab:all}
		\begin{tabular}{|l|c|c|r|c|r|r|r|r|}
			\hline
			Dataset &  Resolution & Color & Num & FPS& Max & Min & Average & Total  \\ \hline\hline
			UCF\_CC\_50 &  different & Grey & 50 & Images & 4543 & 94 & 1279.5 & 63974 \\ \hline
			UCSD & 158 $\times$ 238 & Grey & 2000 & 10 & 46 & 11 & 24.9 & 49885 \\ \hline
			Mall  &  640 $ \times $ 480 & RGB & 2000 & - & 53 & 11 &  31.2 & 62315 \\ \hline
			WorldExpo & 576 $\times$ 720 & RGB & 3980 & 50 & 253 & 1 & 50.2 & 199923 \\ \hline 
		\end{tabular}
	\end{center}
\vspace{-2em}
\end{table*}

\subsection{Evaluation metric}

For crowd counting, the mean absolute error~(MAE) and mean squared error~(MSE) are the two most commonly used evaluation metrics. They are defined as follows:
\begin{equation}
	\mbox{MAE} = \frac{1}{N}\sum_{i=1}^{N} \left|p_{i} - \hat{p}_{i}  \right|, \
	\mbox{MSE} = \sqrt{ \frac{1}{N}\sum_{i=1}^{N} ( p_{i} - \hat{p}_{i} )^2 },
\end{equation}
where $N$ is the total number of frames used for testing, $p_{i}$ and $\hat{p}_{i}$ are the true number and estimated number of people in frame $i$ respectively. As discussed above, $\hat{p}_{i}$ is calculated by summing over the estimated density map over the entire image.

\begin{figure*}[h]
	\centering
	\includegraphics[width = 1.0\textwidth]{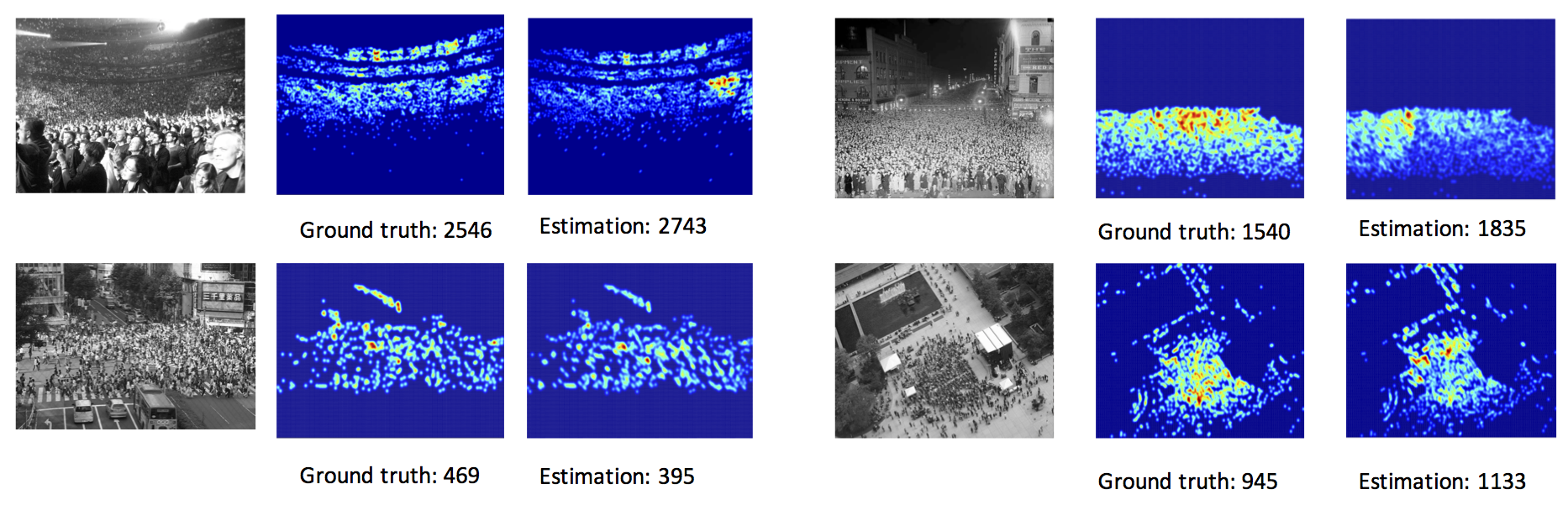}
	\caption{Results for four test images from the UCF\_CC\_50 dataset.  For each example, we show the input image (left), ground-truth density map (middle), and density map obtained by ConvLSTM-nt (right).}
	\label{fig:ucf}
\end{figure*}

\begin{figure*}[t]
	\centering
	\includegraphics[scale = 0.6]{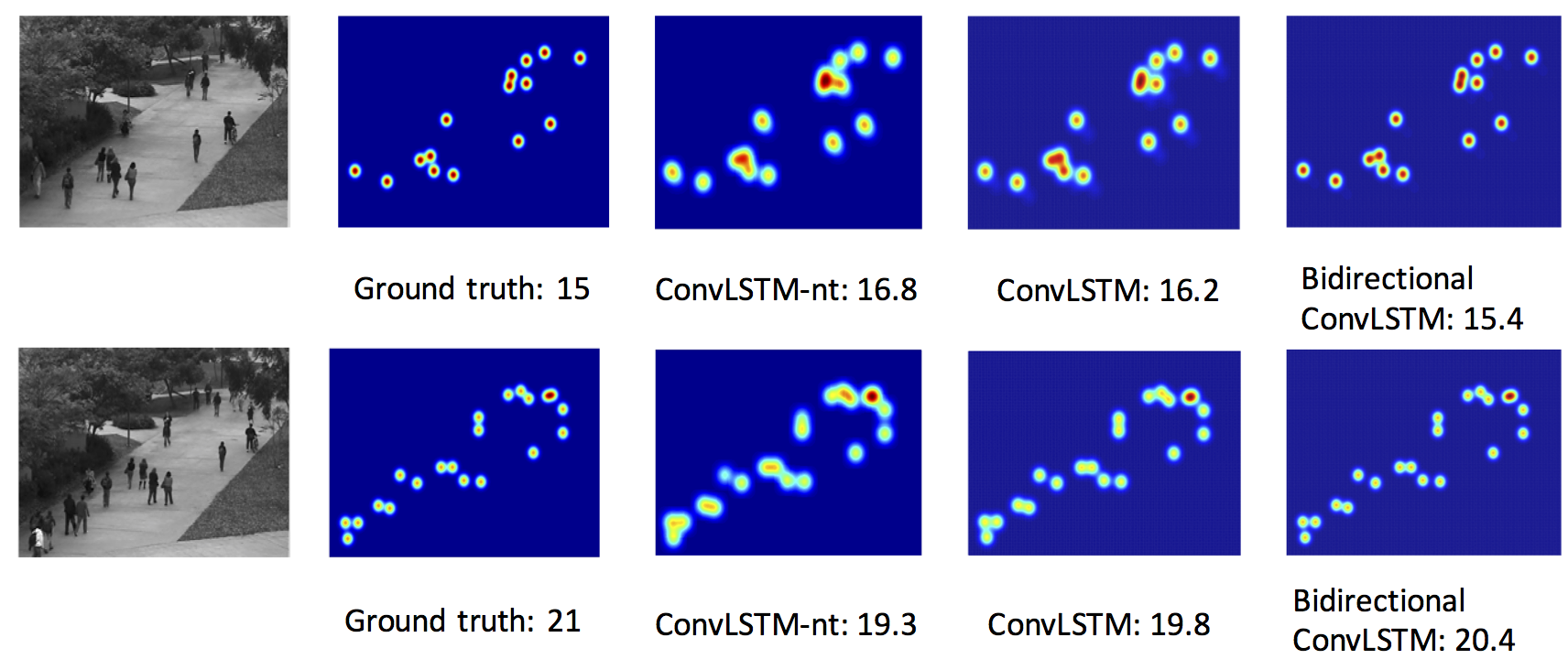}
	\caption{Results for two test video frames from the UCSD dataset.  For each example, we show the input video frame, ground-truth density map, and density maps obtained by the three variants of our method.}
	\label{fig:ucsd}
\end{figure*}

\subsection{UCF\_CC\_50 dataset}

The UCF\_CC\_50 dataset was first introduced by Idress \emph{et al.}~\cite{ucf50}. It is a very challenging dataset because it contains only 50 images of different resolutions, different scenes, and extremely high crowd density. In particular, the number of pedestrians ranges between 94 and 4,543 with an average of 1,280.  Annotations of all the 63,794 people in all 50 images are available in the dataset.  Since the 50 images have no temporal correlation between them, we cannot demonstrate the advantage of exploiting temporal information.  So only the ConvLSTM-nt variant is applied on this dataset.  The goal here is to show that our model can still give very good results for such extremely dense crowd images even though temporal information is not available.

Following the setting in \cite{ucf50}, we split the dataset randomly and perform 5-fold cross validation. To handle different resolutions, we randomly crop patches of size 72 $\times$ 72 from each image for training and testing.  As for the overlapping patches in the test set, we calculate the density at each pixel by averaging the overlapping patches.

\begin{table}[tb!]
	\begin{center}
		\caption{Results of different methods on the UCF\_CC\_50 dataset. It should be noticed that Shang \emph{et al.} \cite{shang2016end} used additional data for training, so it is not fair to compare its result with the others directly.}
		\label{table:ucf}
		\begin{tabular}{ | l | r | r |  }
			\hline
			Method & MAE & MSE \\ \hline\hline
			Head detection \cite{rodriguez2011} & 655.7 & 697.8 \\ \hline
			Density map + MESA \cite{densitymap}  & 493.4 & 487.1 \\ \hline
			Multi-source features   \cite{ucf50}  & 419.5 & 541.6 \\ \hline
			Crowd CNN   \cite{dcnn} & 467.0 & 498.5  \\ \hline
			Multi-column CNN  \cite{mcnn}  & 377.6  & 509.1 \\ \hline
			ConvLSTM-nt &  \bf{284.5} & \bf{297.1}  \\ \hline  \hline
			Shang \emph{et al.} \cite{shang2016end}  & \bf{270.3} & -  \\  \hline
		\end{tabular}
	\end{center}
\vspace{-2em}
\end{table}

We compare our method with six existing methods on the UCF\_CC\_50 dataset.  The results are shown in Table~\ref{table:ucf}. Rodriguez \emph{et al.}~\cite{rodriguez2011} adopted the density map estimation in detection-based methods. Lempitsky \emph{et al.}~\cite{densitymap} extracted 800 dense SIFT features from the input image and learned a density map with the proposed MESA distance (where MESA stands for Maximum Excess over SubArrays).  Idress \emph{et al.}~\cite{ucf50} estimated the crowd count by multi-source features which include SIFT and head detection. The methods proposed by C.~Zhang \emph{et al.}~\cite{dcnn}, Y.~Zhang \emph{et al.}~\cite{mcnn}, and Shang \emph{et al.}~\cite{shang2016end} are all CNN-based methods. Shang \emph{et al.}~\cite{shang2016end} used a model pre-trained on the WorldExpo dataset as initial weights and yielded the best MAE.  However, when considering only methods that do not use additional data for training, our ConvLSTM-nt model achieves the lowest MAE and MSE.

Some results obtained by ConvLSTM-nt are shown in Figure~\ref{fig:ucf}.  Although the images have wide variations in the background and crowd density, ConvLSTM-nt is quite robust in producing reasonable density maps and hence the overall crowd counts.

\subsection{UCSD dataset}

The UCSD dataset~\cite{gpr} contains a 2,000-frame video of pedestrians on a walkway of the UCSD campus captured by a stationary camera. The video was recorded at 10 fps with dimension $238\times158$. The labeled ground truth marks the center of each pedestrian. The ROI and the perspective map are provided in the dataset. 

Using the same setting as in \cite{gpr}, we use frames 601--1,400 as the training data and the remaining 1,200 frames as test data. The provided perspective map is used to adjust the ground-truth density map by setting $ \sigma = 0.3 M(p)$. The values of the pixels outside the ROI are set to zero. 

The results of different methods are shown in Table~\ref{tab:ucsd}. \cite{gpr,ridge,cum} are traditional methods which give the crowd count for the whole image. \cite{densitymap,rf} are density map regression methods using handcrafted features and regression algorithms such as linear regression and random forest regression. Most state-of-the-art methods are based on CNNs~\cite{cnnboosting,dcnn,hydracnn,mcnn}.  Bidirectional ConvLSTM achieves comparable MAE and MSE with these methods. From the results of ConvLSTM-nt, unidirectional ConvLSTM, and bidirectional ConvLSTM , we can draw the conclusion that temporal information can boost the performance for this dataset.

Figure~\ref{fig:ucsd} shows two illustrative examples.  We can see that bidirectional ConvLSTM produces density maps that are closest to the ground truth.  While ConvLSTM-nt can give a rough estimation, ConvLSTM and bidirectional ConvLSTM are more accurate in the fine details.

\begin{table}[h] 
	\begin{center}
		\caption{Results of different methods on the UCSD dataset}
		\label{tab:ucsd}
		\begin{tabular}{|l|r|r|}
			\hline
			Method &  MAE & MSE \\ \hline\hline
			Gaussian process regression \cite{gpr} & 2.24 &  7.97\\ \hline
			Ridge regression \cite{ridge}&  2.25  &  7.82 \\ \hline
			Cumulative attribute regression \cite{cum}& 2.07 & 6.90   \\ \hline
			Density map + MESA \cite{densitymap}  & 1.70   & -   \\   \hline
			Count forest \cite{rf}& 1.60 & 4.40 \\ \hline
			Crowd CNN  \cite{dcnn}&  1.60 & 3.31 \\ \hline
			Multi-column CNN \cite{mcnn} & \bf{1.07}  & \bf{1.35} \\  \hline
			Hydra CNN \cite{hydracnn}  & 1.65 &  -  \\  \hline
			CNN boosting \cite{cnnboosting}  & 1.10  & -   \\   \hline
			ConvLSTM-nt & 1.73 & 3.52 \\  \hline
			ConvLSTM  & 1.30 &  1.79 \\  \hline
			Bidirectional ConvLSTM  & 1.13  & 1.43 \\  \hline 
		\end{tabular}
	\end{center}
\vspace{-2em}
\end{table}

\subsection{Mall dataset}

The Mall dataset was provided by Chen \emph{et al.}~\cite{ridge} for crowd counting. It was captured in a shopping mall using a publicly accessible surveillance camera. This video contains 2,000 annotated frames of moving and stationary pedestrians with more challenging lighting conditions and glass surface reflections. The ROI and the perspective map are also provided in the dataset.

Following the same setting as \cite{ridge}, we use the first 800 frames for training and the remaining 1,200 frames for testing. We perform comparison against Gaussian process regression~\cite{gpr}, ridge regression~\cite{ridge}, cumulative attribute ridge regression~\cite{cum}, and random forest regression~\cite{rf}.  Bidirectional ConvLSTM achieves state-of-the-art performance with respect to both MAE and MSE. The results are shown in Table~\ref{tab:mall}, which also demonstrates the effectiveness of exploiting temporal information.

\begin{table}[tb!] 
	\begin{center}
		\caption{Results of different methods on the Mall dataset}
		\label{tab:mall}
		\begin{tabular}{|l|r|r|}
			\hline
			Method &  MAE & MSE \\ \hline\hline
			Gaussian process regression \cite{gpr} & 3.72 &  20.1\\ \hline
			Ridge regression \cite{ridge}&  3.59  &  19.0 \\ \hline
			Cumulative attribute regression \cite{cum}& 3.43 & 17.7   \\ \hline
			Count forest \cite{rf}& 2.50 & 10.0 \\ \hline
			ConvLSTM-nt  &  2.53 & 11.2 \\  \hline
			ConvLSTM  & 2.24 &  8.5 \\  \hline
			Bidirectional ConvLSTM  & \bf{2.10}  & \bf{7.6} \\  \hline 
		\end{tabular}
	\end{center}
\vspace{-2em}
\end{table}    

\subsection{WorldExpo dataset}

The WorldExpo dataset was introduced by C.~Zhang \emph{et al.}~\cite{dcnn,zhang2016data}.  This dataset contains 1,132 annotated video sequences captured by 108 surveillance cameras, all from the 2010 Shanghai World Expo.  The annotations of 199,923 pedestrians in 3,980 frames include the location of the center of each human head.  The test set contains five separate video sequences each of which has 120 annotated frames.  The regions of interest (ROIs) are provided for these five test scenes.  The perspective maps are also provided.

\begin{table*}[h]
	\begin{center}
		\caption{Results of different methods on the WorldExpo dataset}
		\label{tab:worldexpo}
		\begin{tabular}{ | l | r | r | r | r | r | r | }
			\hline
			Method  &  Scene 1 & Scene 2 & Scene 3 & Scene 4 & Scene 5 & Average \\ \hline\hline
			LBP features + ridge regression & 13.6  & 59.8 & 37.1 & 21.8 & 23.4& 31.0 \\ \hline
			Deep CNN \cite{dcnn}& 9.8  & \bf{14.1} & 14.3 & 22.2 & 3.7 & 12.9 \\ \hline
			Multi-column CNN \cite{mcnn} & \bf{3.4}  & 20.6 & \bf{12.9} & \bf{13.0} & 8.1 & 11.6 \\ \hline
			ConvLSTM-nt & 8.6  & 16.9 & 14.6 & 15.4 & 4.0 & 11.9 \\ \hline
			ConvLSTM  & 7.1  & 15.2 & 15.2 & 13.9 & 3.5 & 10.9 \\ \hline
			Bidirectional ConvLSTM  & 6.8  & 14.5 & 14.9 & 13.5 & \bf{3.1} & \bf{10.6} \\ \hline
		\end{tabular}
	\end{center}
\vspace{-2em}
\end{table*}

\begin{figure}
	\centering
	\includegraphics[width = 0.45\textwidth]{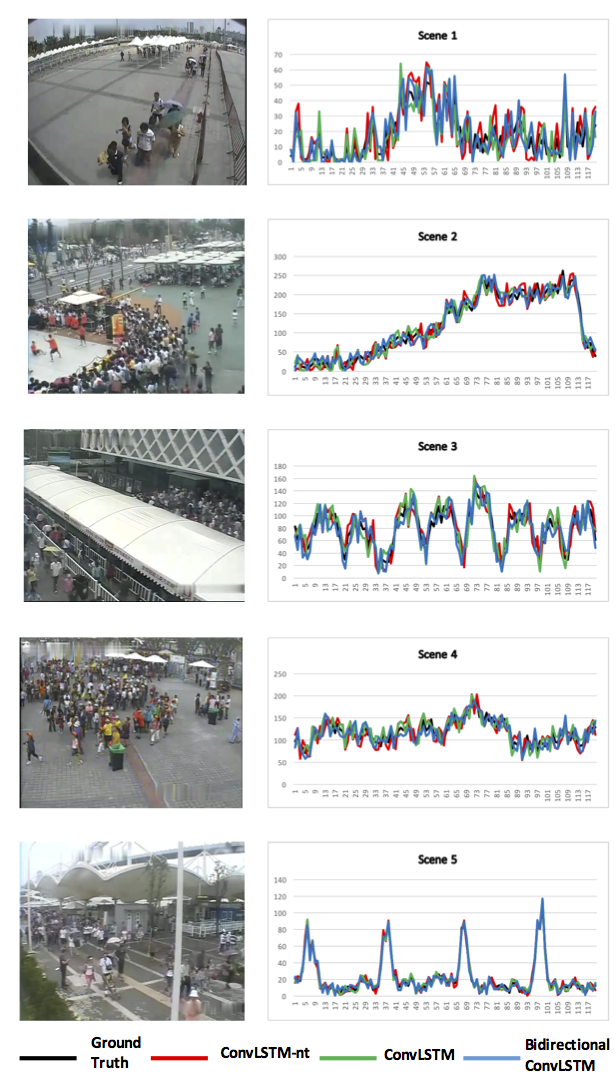}
	\caption{Density map estimation examples from the WorldExpo dataset (best viewed in color). In each row, the left one shows one video frame from the test scene and the right one shows the estimation results of that scene, where the $x$-axis represents the frame index and the $y$-axis represents the crowd count.}
	\label{fig:worldexpo}
	\vspace{-1em}
\end{figure}

For fair comparison, we follow the work of the multi-column CNN to generate the density map according to the perspective map with the relation $ \delta  = 0.2 ∗ M (x)$, where $M (x) $ denotes the number of pixels in the image representing one square meter at location $x$. Table \ref{tab:worldexpo} compares our model and its variants with the state-of-the-art methods. We use MAE as the evaluation metric, as suggested by the author of \cite{dcnn}.  On average, bidirectional ConvLSTM achieves the lowest MAE.  It also gives the best result for scene 5.

We show the estimation results for the five test scenes obtained by our models in Figure~\ref{fig:worldexpo}.  The crowd count curves are shown in different colors for the ground truth (black) and the estimation results of ConvLSTM-nt (red), ConvLSTM (green), and bidirectional ConvLSTM (blue).  We note that the five scenes differ significantly in the scene type, crowd density, and change in crowd count over time.

\begin{figure}
	\centering
	\includegraphics[width = 0.4\textwidth]{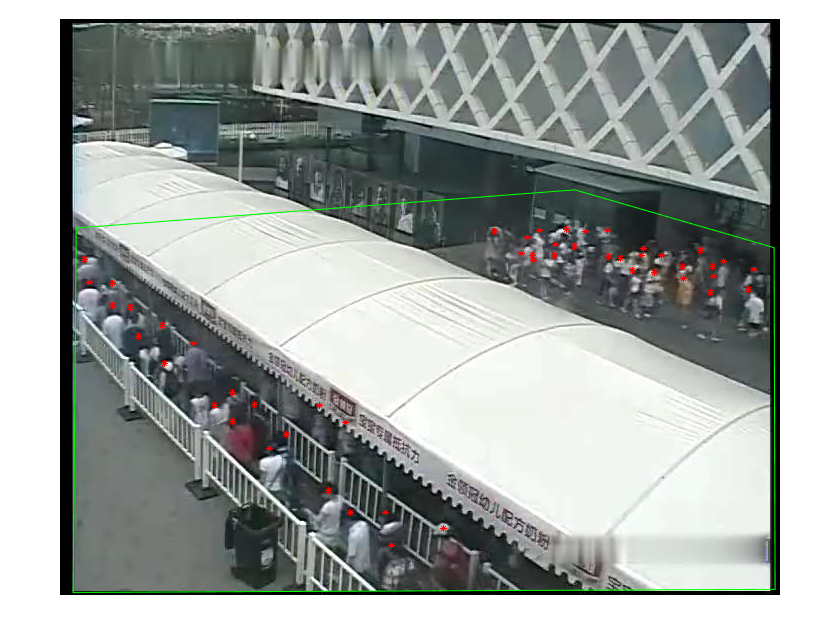}
	\caption{A video frame from test scene 3 of the WorldExpo dataset.  The region outlined in green indicates the ROI and the red dots mark the positions of the heads.}
	\label{fig:scene3}
\end{figure}

From Table~\ref{tab:worldexpo} and Figure~\ref{fig:worldexpo}, we can see that bidirectional ConvLSTM outperforms ConvLSTM and ConvLSTM outperforms ConvLSTM-nt in most cases (scene 1,2,4,5), which gives evidence to the effectiveness of incorporating temporal information for crowd counting.  As for scene 3, a closer investigation reveals a potential problem with the labels provided in this test scene.  Figure~\ref{fig:scene3} illustrates the problem.  There are in fact many people walking under the white ceiling of the covered walkway as we can see their moving legs clearly when playing the video, but only two red dots are provided in the frame because the heads of most of the people there are hidden.  Spatiotemporal models tend to count them since motion is detected when exploiting the temporal information, but unfortunately they are not annotated in the provided labels.

\subsection{Transfer learning experiments}

\begin{figure}
	\centering
	\includegraphics[width = 0.45\textwidth]{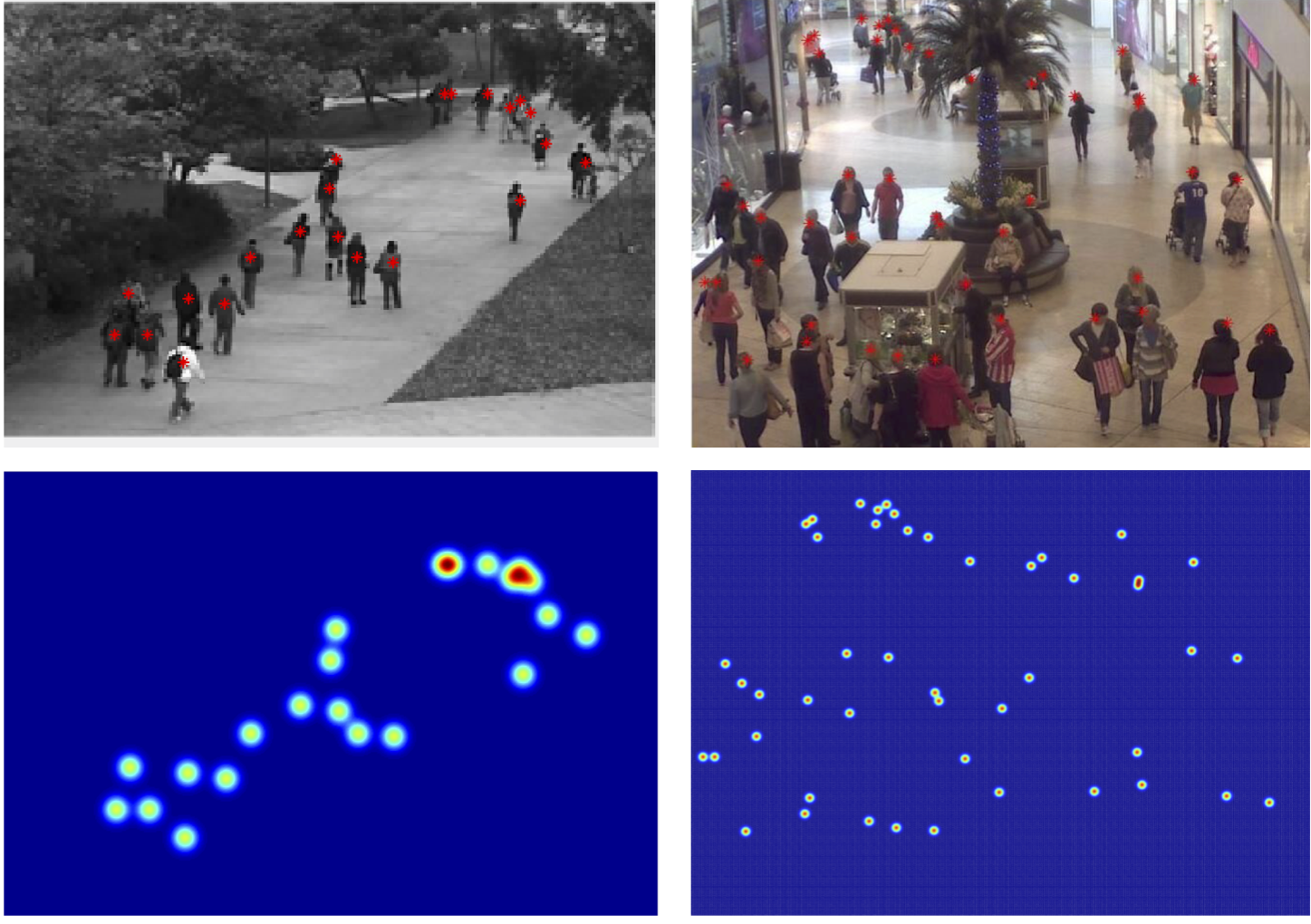}
	\caption{The UCSD and Mall datasets used for transfer learning experiments. Left column: UCSD dataset; right column: Mall dataset. Upper row: input images with annotations; lower row: density maps. }
	\label{fig:transfer}
\end{figure}

To demonstrate the generalization capability of our model, we conduct some experiments in the transfer learning setting. Specifically, we compare with some previous methods that have also been evaluated in the transfer learning setting using the UCSD and Mall datasets, which were both captured using stationary cameras.  As shown in Figure~\ref{fig:transfer}, the two datasets are quite different in terms of the scene type (outdoor for UCSD but indoor for Mall), crowd density, frame rate, and camera angle, among others.

We consider two transfer learning tasks by using one dataset as the source domain and the other one as the target domain.  For each task, 800 frames are used for training the model and 50 frames of the other dataset are used as the adaptation set. Following the same setting as \cite{change2013semi,yu2005learning,liu2015bayesian}, we use MAE as the evaluation metric. Table \ref{tab:transfer} presents the results for different methods on the two transfer learning tasks. Bidirectional ConvLSTM achieves state-of-the-art performance in both transfer learning tasks. We note that the performance of our method in the transfer learning setting is even better than many approaches tested on the standard, non-transfer-learning setting. For instance, with 800 frames of the UCSD dataset for training and 50 frames of the Mall dataset for adaptation, bidirectional ConvLSTM can achieve an MAE of 2.63, which outperforms many algorithms using 800 frames of the Mall dataset for training, according to Table~\ref{tab:mall}. We can draw the conclusion that bidirectional ConvLSTM has good generalization capability. Once trained on one dataset, the learning experience can be transferred easily to a new dataset which consists of only very few video frames for adaptation. 

\begin{table}[h!]
	\begin{center}
		\caption{Results of transfer learning across datasets with MAE as evaluation metric. FA: feature alignment; HGP: hierarchical Gaussian process; GPA: Gaussian process adaptation; GPTL: Gaussian process with transfer learning.}
		\label{tab:transfer}
		\begin{tabular}{|l|c|c|}
			\hline
			&  UCSD to & Mall to \\
			Method  &  Mall & UCSD \\ \hline\hline
			FA \cite{change2013semi}  &  7.47 & 4.44 \\ \hline
			HGP \cite{yu2005learning}  &  4.36  &  3.32   \\  \hline
			GPA \cite{liu2015bayesian}  &  4.18 &  2.79  \\ \hline
			GPTL \cite{liu2015bayesian}  &  3.55 &  2.91  \\ \hline
			Bidirectional ConvLSTM & \bf{2.63} &  \bf{1.82}  \\ \hline
			
		\end{tabular}
	\end{center}
	\vspace{-2em}
\end{table}      
%



\section{Conclusion}

In this paper, we have pursued the direction of spatiotemporal modeling for improving crowd counting in videos.  By jointly capturing both spatial and temporal dependencies, we overcome a major limitation of the recent CNN-based crowd counting methods and advance the state of the art.  Specifically, our models outperform existing crowd counting methods on the UCF\_CC\_50 dataset, Mall dataset, and WorldExpo dataset, and achieve comparable results on the UCSD dataset.  The superior result on the UCF\_CC\_50 dataset shows that our model can still perform well on extremely dense crowd images even when temporal information is not available.  As for the other three datasets, the results show that explicitly exploiting temporal information has a clear advantage.  Finally, the last set of experiments shows that our model is robust under the transfer learning setting to generalize from previous learning experience.

In the future, we are going to extend our model to deal with the active learning setting for crowd counting. We will output an additional confidence map and actively query the labeler to label only the less confident regions, which would greatly alleviate the expensive labeling effort for crowd counting in videos. 

\section{Acknowledgement}
This research has been supported by General Research Fund 16207316 from the Research Grants Council of Hong Kong and Innovation and Technology Fund ITS/170/15FP from the Innovation and Technology Commission of Hong Kong.

{\small
\bibliographystyle{ieee}
\bibliography{egbib}
}

\end{document}